\documentclass[conference]{IEEEtran}

\usepackage{cite}
\usepackage{amssymb,amsmath,amsfonts,hyperref}
\usepackage{graphicx}
\usepackage[utf8]{inputenc}

\begin{document}

\title{An Attention Free Long Short-Term Memory for Time Series Forecasting} 

\author{\IEEEauthorblockN{Hugo Inzirillo}
\IEEEauthorblockA{\textit{CREST - Institut Polytechnique de Paris}\\
\textit{Simons}\\
Email: hugo@simons.finance}
\and
\IEEEauthorblockN{Ludovic De Villelongue}
\IEEEauthorblockA{\textit{Université Paris Dauphine-PSL}\\
\textit{Simons}\\
Email: ludovic.de-villelongue@dauphine.eu}
\\
}

\maketitle

\begin{abstract} 
Deep learning is playing an increasingly important role in time series analysis. We focused on time series forecasting using attention free mechanism, a more efficient framework, and proposed a new architecture for time series prediction for which linear models seem to be unable to capture the time dependence. We proposed an architecture built using attention free LSTM layers that overcome linear models for conditional variance prediction. Our findings confirm the validity of our model, which also allowed to improve the prediction capacity of a LSTM, while improving the efficiency of the learning task.
\end{abstract}

\IEEEpeerreviewmaketitle

\bibliographystyle{IEEEtran}

\section{Introduction}

 Recently we have seen a strong interest in the development of models based on attention mechanisms, for encoder decoder based models. More recently, some researchers have focused on the development of more efficient mechanisms, and developed some alternatives such as the attention free mechanism \cite{attention_free_transformer} an efficient variant of Transformers \cite{attention_is_all_u_need}. The initial problem was to improve the performance of the encoder-decoder model. The one behind the attention mechanism was to allow the decoder to use the most relevant parts. However, the complex structure of these transformers does not seem to significantly improve the prediction when dealing with time series \cite{dlinear}. In this paper we have retained the usefulness of the attention mechanism and more particularly its capacity to use the most relevant parts of the input sequence. We have combined this strength with the power of a LSTM \cite{lstm} to design a new framework for time series prediction for assets for which linear models are no longer sufficient to model the conditional volatility. 
 
 Digital assets became popular for their high volatility, typically where linear models may have difficulties on predictiont task. In this paper we focus on the modeling of volatility by feeding the outputs of a GARCH model and adjust the prediction using an Attention Free Long Short-Term Memory (AF-LSTM). The results are promising on various assets and make the AF-LSTM Layer a more efficient predicator for time series.


\section{Deep Volatility}

The assumption made about the errors of classical linear models is that they are homoscedastic, they follow a centered distribution with a standard deviation $\sigma$. This assumption is not realistic in financial applications. In fact, in practice, we find that the errors are heteroscedastic and this is more or less strong depending on the asset class. In other words, the variance is time dependent. When we look at certain asset classes, we observe volatility clusters and this is even stronger when we talk about digital assets. This characteristic is typical of conditional heteroscedasticity\cite{BOLLERSLEV1986307}.

\subsection{Autoregressive conditional hererocesdastic models}
ARCH is a univariate and nonlinear model in which volatility is estimated with the square of past returns.
Despite its appealing features, such as simplicity, nonlinearity, easiness, and adjustment for forecast, the ARCH model has certain drawbacks:
\begin{enumerate}
\item equal response to positive and negative shocks
\item strong assumptions such as restrictions on parameters
\item possible misprediction due to slow adjustments to large movements
\end{enumerate}

\subsection{GARCH Model}
GARCH model \cite{BOLLERSLEV1986307} is also an approach to attempt to
measure the volatility of certain financial assets. GARCH models has been seen for
decades as one of the leading tools in terms of estimating the volatility of the financial returns.
Three main reasons have made GARCH models so popular: 

\begin{enumerate}
    \item the variance of residuals is auto-correlated
    \item it allows to manage volatility clusters
    \item it represents the conditional variance for each step of time
\end{enumerate}

A GARCH(P,Q) is defined as:

\begin{equation}
    \sigma_{t}^2=\omega + \sum_{p=1}^{P} \alpha_p \epsilon_{t-p}^2  + \sum_{q=1}^{Q} \beta_q \sigma_{t-q}^{2} 
\end{equation}

where $\epsilon_t = \sigma_t \mu_t$, and $\mu_t \overset{i.i.d}{\sim} \mathcal{N}(0,1)$. If  $\alpha + \beta < 1 $ the volatility fluctuates around the unconditional variance and is given by:

$$
\sigma^2 := Var(r_t) = \frac{\omega}{1- \alpha - \beta} 
$$

\subsection{GJR-GARCH Model}
The Glosten-Jagannathan-Runkle GARCH (GJR-GARCH) \cite{GJRGARCH} is a variant of a GARCH model, the underlying assumption is that negative shocks in the previous period have a stronger impact on the variance than positive shocks.
The GJR-GARCH(P,Q) is defined as:

\begin{equation}
    \sigma_{t}^2=\omega + \sum_{p=1}^{P} \alpha_p \epsilon_{t-p}^2  \gamma_p \epsilon_{t-p}^2 \mathbf{1}_{\{\epsilon_{t-p}<0\}} + \sum_{q=1}^{Q} \beta_q \sigma_{t-q}^{2} 
\end{equation}

where $\epsilon_t = \sigma_t \mu_t$, and $\mu_t \overset{i.i.d}{\sim} \mathcal{N}(0,1)$. $\gamma_p$ equals to 0 implies that there is no asymmetric shock on the time series. If  $\alpha + \frac{\gamma}{2} + \beta < 1 $ the volatility fluctuates around the unconditional variance, given by:

$$
\sigma_{GJR}^2 := Var(r_t) = \frac{\omega}{1- \alpha - \frac{\gamma}{2} - \beta} 
$$

\medskip

Deep learning models are proposed to estimate the conditional volatility of cryptocurrencies. Recurring neural networks (RNNs) are built to improve GARCH predictions. For this purpose, a LSTM using sequences of  past values of volatility is taken as starting point and updated with embeddings corresponding to the GARCH predicted volatility.

\subsection{LSTM}

The LSTM framework \cite{lstm} are RNNs with gated mechanisms designed to avoid vanishing gradient. The blocks of LSTM contain 3 non-linear gates that make it smarter than a classical neuron, as well as a memory for sequences. The 3 types of non-linear gates include : 
\begin{itemize}
    \item Input Gate ($i_t$) : decides which values from the input are used to update the memory
    \item Forget Gate ($f_t$) : handles what information to throw away from the block
    \item Output Gate ($o_t$): handles what will be in output based on input and memory gate.
\end{itemize}

During the training step, each iteration provides an update of the model weights proportional to the partial derivative and in some cases the gradient may be vanishingly small and weights may not be updated.
The LSTM networks is defined in Figure \ref{fig:lstm_cell} as:

\begin{figure}[h!]
    \centering
\includegraphics[width=1\columnwidth]{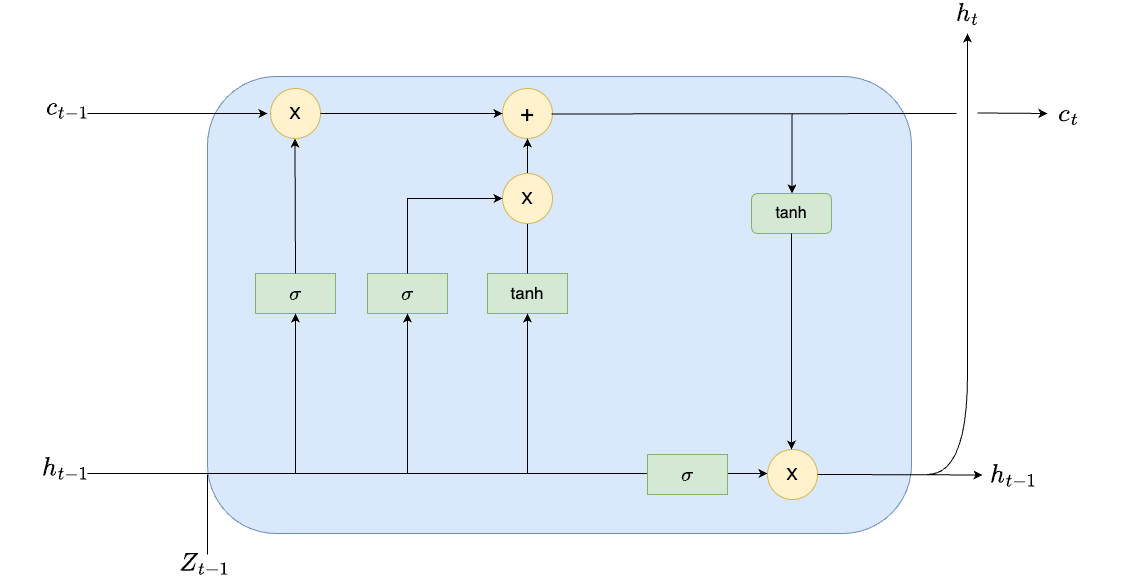}
\caption{LSTM Cell}
\label{fig:lstm_cell}
\end{figure}

\begin{equation}
\label{eq:lstm}
\begin{split}
    f_{t} & = \sigma(W_{f}[h_{t-1},z_{t-1}]+ b_{f}),\\
    i_{t} & = \sigma(W_{i}[h_{t-1},z_{t-1}]+ b_{i}),\\
    {\tilde c}_{t} & = {\tanh (W_{c}[h_{t-1},z_{t-1}]+ b_{c})},\\
    c_{t} & = f_{t} \odot c_{t-1}+i_{t} \odot {\tilde c}_{t},\\
    o_{t} & = \sigma(W_{o}[h_{t-1},z_{t-1}]+ b_{o}),\\ 
    h_{t} & = o_{t} \odot {\tanh(c_{t}).} 
\end{split}
\end{equation}
where $ \Theta \triangleq \{W_f, W_i, W_c, W_o\}$ are the weights of the model and  $z_{t-1}$ is the input vector at time t. At each time step, the memory cell takes the current input $z_{t-1}$, the previous hidden state $h_{t-1}$ and the previous memory state $c_{t-1}$. 

The goal is to estimate $\hat{y_t} \triangleq$ $f(h_t;\theta)$. In this case it is equivalent to estimate:

$$\mathbb{E}[\sigma_{t}^2|\mathcal{F}_{t-1}] = \mathbb{E}[\sigma_{t}^2|(\sigma_{t-1}^2,..., \sigma_{t-q}^2),(\epsilon_{t-1}^2,..., \epsilon_{t-p}^2) ]. $$

The loss function given by $L:Y \times Y \rightarrow {\mathbb{R}_+} $ measures the distance between two outputs. In this case, the $l_2-loss$ is used. It is given by:

\begin{equation}
    \begin{split}
            L(y,f(z;\theta)) & = (y-f(z_{0:T-1};\theta))^2,\\
            & = \frac{1}{T} \sum_{t=1}^{T} (y_t- f(z_{t-1};\theta))^{2}.
    \end{split}
\end{equation}

The number of parameters of each layer in the model should be carefully selected to handle overfitting or underfitting situations.

\subsection{Deep GARCH}
With the expanding need to improve volatility modeling precision, GARCH models have been a redundant concern in finance and economic literature. However, the relation between the conditional variances being non linear, these models have found their limits. Some researcher works focusing on introducing non linearity in these models rely on neural networks to introduce non-linearity into classical models \cite{NIKOLAEV2013501,KRISTJANPOLLER20157245}. Kim and Won \cite{KIM201825} mixed the LSTM with multiple GARCH-type models to introduce non linearity \cite{GarchANN}. In this paper we seek to improve the predictive ability of GARCH models. We do not question the estimation of the model but we try to correct the prediction made at each time step. To do this, we recover the conditional variances obtained after a first estimation made by the GARCH model and then we apply filters and linear transformations made possible by the use of several AF-LSTM layers.

\subsection{Attention Free Block}
The transformer architecture has made it possible to develop new models capable of being trained on large dataset while being much better than recurrent neural networks such as LSTM. The Attention Free Transformer \cite{attention_free_transformer} introduces the attention free block, an alternative to attention block \cite{attention_is_all_u_need},  which eliminates the need for dot product self attention. The transformer is an efficient tool to capture the long term dependency. Just like the transformer, the AF Block includes interactions between queries, keys and values. The difference is that the AF block firstly combines key and value together with a set of learned position biases described in \cite{attention_free_transformer}. Then the query is combined with the context vector. Let Q, K, V denote the query, key and value, respectively.

\begin{equation}
    \begin{split}
        Y_t & =  \sigma_q(Q_t) \odot \frac{\sum_{t^{'}=1}^T exp(K_{t^{'}}+w_{t,t^{'}}) \odot V_{t^{'}} }{\sum_{t^{'}=1}^T exp(K_{t^{'}}+w_{t,t^{'}}))},\\
         &= \sigma_q(Q_t) \odot  \sum_{t^{'}=1}^T (\text{Sofmax}(K)\odot V)_{t{'}}.
    \end{split}
\end{equation}

where $Q=ZW_q$, $K=ZW_k$ and $V=ZW_v$. The activation function $\sigma_q$  is the sigmoid function and $w_t \in \mathbb{R}^{T\times T}$ is a learned matrix of pair-wise position biases.

\subsection{Attention-free LSTM}
As explained in the models described above, the expectation of volatility is conditioned by past values. However some of these variables may at some point not be relevant for the prediction. To reproduce this observable phenomenon, we added an Attention Free Block to build an Attention-free LSTM layer (AF-LSTM). The Attention-free LSTM layer processing step can be split in several parts. The input at each time step will firstly be processed by an Attention Free layer defined as: 

\begin{equation}
    \Tilde{Z}_{t-1}^{(l)} = AF^{(l)}_1(W_{1}^{(l)};z_{t-1}),
\end{equation}

where $Z_t \in \mathbb{R}^q$, $q$ the sequence length of our input and $AF^{(l)}_{1}$ is the Attention Free of the $l-th$ layer:

\begin{equation}
\label{eq:af_mechanism}
\begin{split}
    x_{t-1}^{(l)} & = \text{Concat}(W_{x}^{(l)}Z_{t-1}^{(l)} + b_x^{(l)}),\\
    Q_{t-1}^{(l)} & = W_{q}^{(l)}  x_{t-1}^{(l)},\\
    K_{t-1}^{(l)} &=  W_{k}^{(l)}  x_{t-1}^{(l)},\\
    V_{t-1}^{(l)} & = W_{v}^{(l)}  x_{t-1}^{(l)}, \\
    \eta_{t-1}^{(l)} &=  \sum_{t=1}^T \text{Softmax}( K_{t-1}^{(l)} )\odot V_{t-1}^{(l)},\\
    \Tilde{Z}_{t-1}^{(l)} &= \sigma( Q_{t-1}^{(l)}) \odot \eta_{t-1}^{(l)}.\\
\end{split}
\end{equation}

we apply the \textit{Layer Normalization (LN)} function \cite{ba2016layer}

$$
LN(x;\psi;\phi)= \psi \frac{(x-\mu_x)}{\sigma_x}+\phi,
$$

and filter our $\Tilde{Z}_{t-1}^{(l)}$ using the \textit{ReLU} activation function, $ReLU(x)=max(0,x)$.

Hence we have:

\begin{equation}
    \Tilde{z}_{t-1}^{(l)} = ReLU(LayerNorm(\Tilde{Z}_{t-1}^{(l)})),
\end{equation}

\begin{figure}[h!]
    \centering
\includegraphics[width=1\columnwidth]{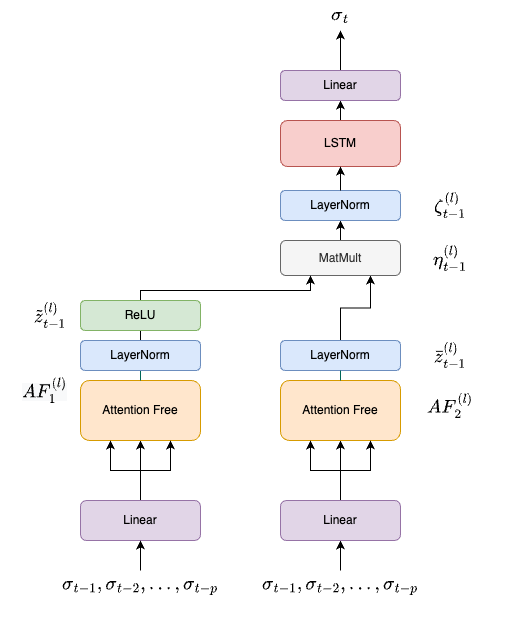}
\caption{Attention-Free LSTM Layer}
\label{fig:af_lstm_layer}
\end{figure}

On the right side of \ref{fig:af_lstm_layer} the input is the same as the left size. We denote $\Bar{Z}_{t-1}$ the output of the attention-free mechanism $AF_{2}^{(l)}$.  $\Bar{Z}_{t-1}$ will resume the information of the input tensor hence it will be multiplied by the ouput of the left side to only select the observations that have a positive impact on the prediction.

\begin{equation}
\begin{split}
    \Bar{Z}_{t-1}^{(l)} & = AF_{2}^{(l)}(W_{2}^{(l)};Z_{t-1}),\\
    \Bar{z}_{t-1}^{(l)}  & = LN(\Bar{Z}_{t-1};\psi^{(l)};\phi^{(l)}).
\end{split}
\end{equation}

The output from the two channels will be multiplied to apply the filter on our input sequence.

\begin{equation}
    \begin{split}
        \eta_{t-1}^{(l)} &= ( \Tilde{z}_{t-1}^{(l)} \odot \Bar{z}_{t-1}^{(l)}),\\
    \zeta_{t-1}^{(l)} &= LN( \eta^{(l)}_{t-1};\psi^{(l)};\phi^{(l)}).
    \end{split}
\end{equation}

$\zeta_{t-1}^{(l)}$ will be the input of a parametrized function $g_{w}^{(l)}$ such as:

\begin{equation}
    h_{t}^{(l)} = g_{w}^{(l)}(\zeta_{t-1}^{(l)}),
\end{equation}

in Fig \ref{fig:af_lstm_layer} $g_{w}^{(l)}(.)$ is a LSTM defined in Fig \ref{fig:lstm_cell}. Using the output of $g_{w}^{(l)}(.)$ we can estimate the output of the $l-th$ layer $\sigma_{t}^{(l)}$ given by:

\begin{equation}
    \sigma_{t-1}^{(l_{final})} = W_{y}^{(l_{final})}h_{t-1}^{(l_{final})} + b_{y}^{(l_{final})}, \quad l:=\{1,...,l_{final}\}.
\end{equation}

The ouput of our network is defined as $\sigma_t$ which is a proxy of the 5d rolling volatility.

\section{Methodology}
\subsection{Empirical Risk Minimization}
Considering the training data $\mathcal{D} \triangleq \{Z^{(i)}\}_{i=1}^m \in \mathcal{Z}^m $ and a function $f:\mathcal{Z}->\mathcal{Y}$, the ERM is defined by:

\begin{equation}
\hat{\mathcal{R}}_{\mathcal{D}}(f):=  \mathcal{L}(f,Z^{(i)}).
\end{equation}

Lets define $\mathcal{A}_{erm}$ as an empirical risk minimization algorithm. Hence, given a set of $\mathcal{F}$, the objective function is:

\begin{equation}
   A_{erm} (\mathcal{D})   = \underset{f \in \mathcal{F}}{\text{arg min}}\quad \hat{\mathcal{R}}_{\mathcal{D}}(f), 
\end{equation}

where f is the residual mean squared error (RMSE):
\begin{equation}
    RMSE = \sqrt{\frac{\sum_{t=1}^{n} \left( Y_t- \hat{Y}_t \right)^2}{n}}.
\end{equation}

\subsection{Data}
The data retrieved are the close prices of a single cryptocurrency asset from inception.
 The log returns of this time series are calculated and fed to the GARCH model. The distribution chosen for the standardized residuals is a normal one. The lag parameters are defined to perform a first order GARCH model. Once the model is fitted, the forecasted log returns are given by the following formula: 
 \begin{equation}
 \begin{split}
& r_t=\sigma_t \epsilon_t, \\
& \sigma_t^{2}= \omega + \alpha r^{2}_{t-1} +\beta \sigma^{2}_{t-1}.
\end{split}
\end{equation}

\medskip

Then, the volatility calculated on a five period rolling window is compared  with the forecasted volatility obtained by dividing the GARCH forecasted log returns with $\epsilon_t$. The volatility at time step t is calculated by using the moving average of the previous 4 time steps' returns (from time step t-5 to time step t-1).The train and test sets are composed of real volatilities and volatilities forecasted by the GARCH model. They are split according to the 80/20 rule on a five period window and plugged to the LSTM and AF-LSTM. Once the train and test sets are created, two scalers are initialized on $x$ and $y$ to scale and translate them to a (0,1) range. Afterwards, the train set for $x$ and $y$ are fitted and transformed, and the test set for $x$ and $y$  only transformed. The fit method is calculating the mean and variance of each of the features present in the data. The transform method is transforming all the features using the respective mean and variance. The parameters learned by the model using the training data will help to transform the test data. Once it is done, the train and test sets arrays are converted to tensors and inserted into dictionaries with the scalers.\\
\medskip

The LSTM network is fed with the following parameters : 
\begin{itemize}
    \item \textit{Input size}: 2 
    (number of expected features in the input)
    \item \textit{Hidden size}: 64
     (number of features in hidden state h)
\end{itemize}
\medskip

The AF-LSTM network is fed with the following parameters:

\begin{itemize}
    \item \textit{Input size}: 2 
    (number of expected features in the input)
    \item \textit{Hidden size}: 64
     (number of features in hidden state h)
    \item \textit{Maximum sequence length}: 1000
    (maximum number of timesteps to be fed in)
    \item \textit{Dim}: 2 
    (embedding dimension of the keys, queries and values)
    \item \textit{Hidden size} : 64 (hidden dimension used inside the Attention Free Transformer)
\end{itemize}

After creation of the LSTM and AF-LSTM networks with an Adam optimizer, the mean squared error loss function is computed on 1000 epochs with a learning rate of 0.001. Once the predictions are obtained, the scalings of the predicted and actual volatilities are inversed to get back to true volatility values. For comparison purposes, the results are transformed to single arrays. The root mean squared errors (RMSE) for the train and test sets can then be calculated. Finally, the actual volatilities and predicted volatilities for the train and test sets are plotted using the two models. 

\subsection{Estimation}

\begin{figure}[h!]
\centering
\includegraphics[width=0.5\textwidth]{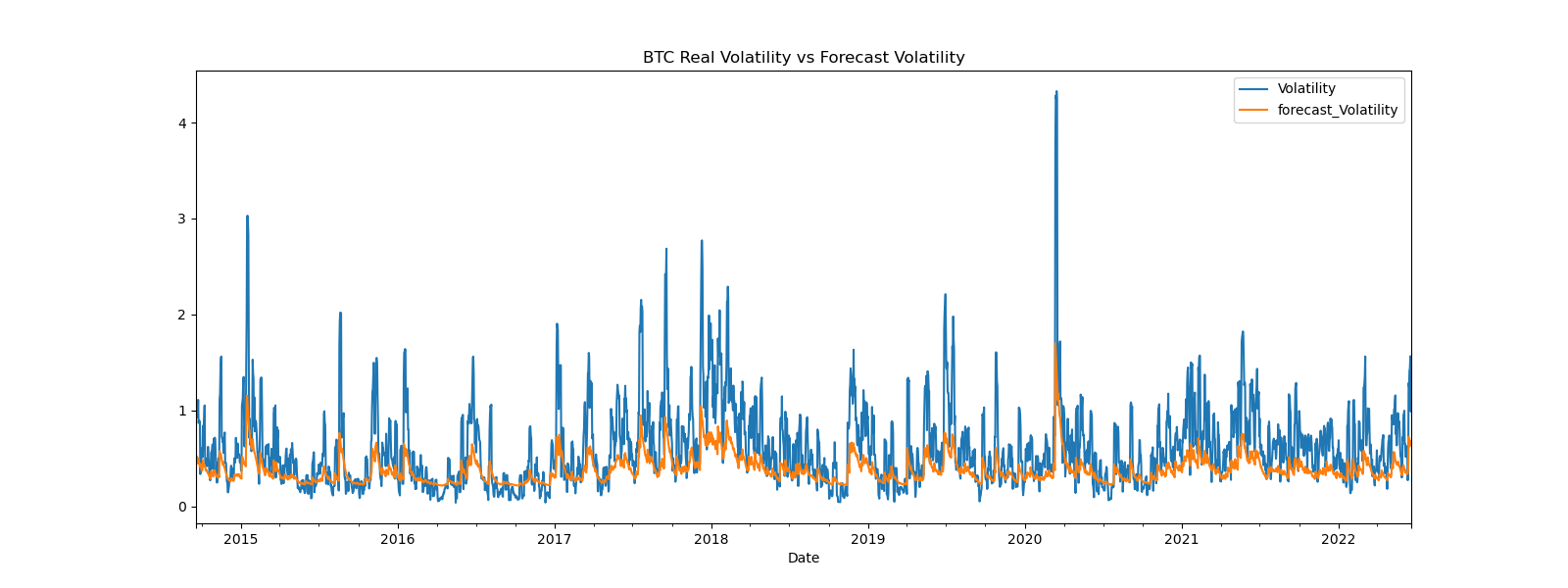}
\caption{Actual Volatility  vs Forecasted Volatility by GARCH(1,1)}
\label{fig:GARCH_actual_pred_vol}
\end{figure}

A scalability matter is encountered when comparing the GARCH forecasted volatility with the actual rolling volatility in Figure \ref{fig:GARCH_actual_pred_vol}. The GARCH (1,1) model is thus a weak learner for such a time series.

\begin{figure}[h!]
\centering
\includegraphics[scale=0.4]{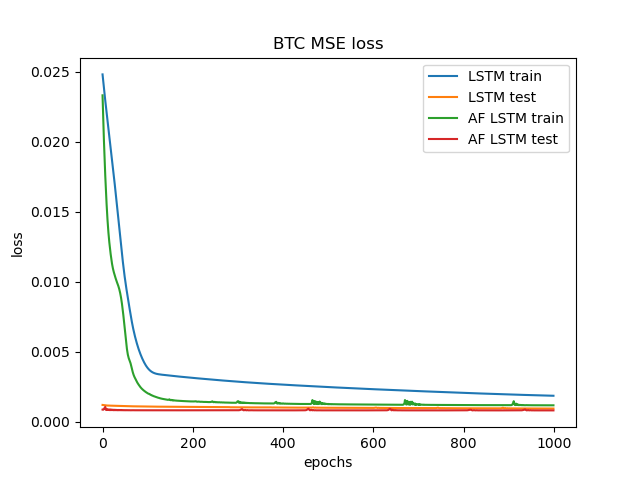}
\caption{ LSTM and AF-LSTM Loss Functions for Train and Test Sets}
\label{fig:Loss_Func}
\end{figure}

\begin{figure}[h!]
\centering
\includegraphics[scale=0.4]{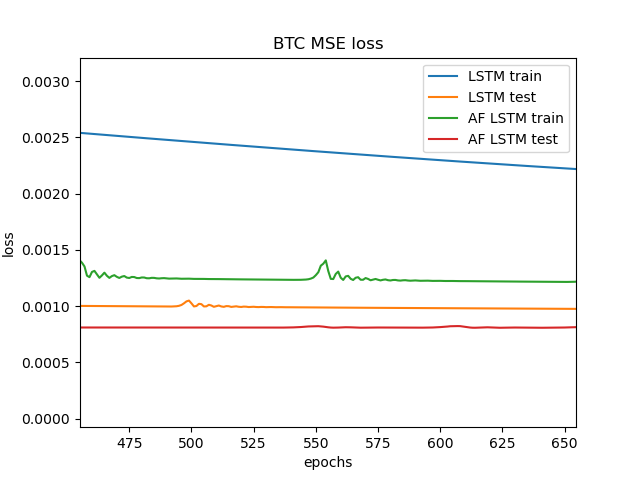}
\caption{LSTM and AF-LSTM Loss Functions Zoomed for Train and Test Sets}
\label{fig:Loss_Func_Zoom}
\end{figure}

The loss function in Figure \ref{fig:Loss_Func} is enlightning the better performance of the AF-LSTM in taking information into account. On the one hand, regarding the train set, the loss of the AF-LSTM is lower than the LSTM's loss on all epochs. On the other hand, regarding the test set, both losses are bottommost and are decreasing at a truly slight rate. On a smaller interval, it is clearly noticeable that the AF-LSTM has a lower loss than the LSTM on the test set, as enlightened in Figure \ref{fig:Loss_Func_Zoom}. It is the case for the integrality of the epochs.

\begin{figure}[h!]
\centering
\includegraphics[scale=0.4]{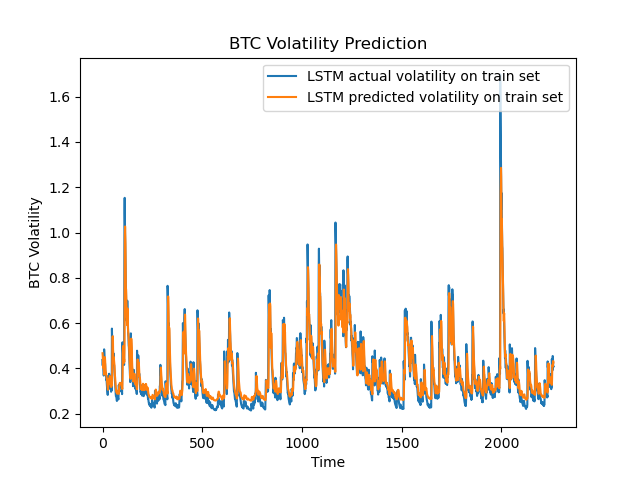}
\caption{Actual Volatility vs LSTM Forecasted Volatility for Train Set}
\label{fig:LSTM_train}
\end{figure}

\begin{figure}[h!]
    \centering
\includegraphics[scale=0.4]{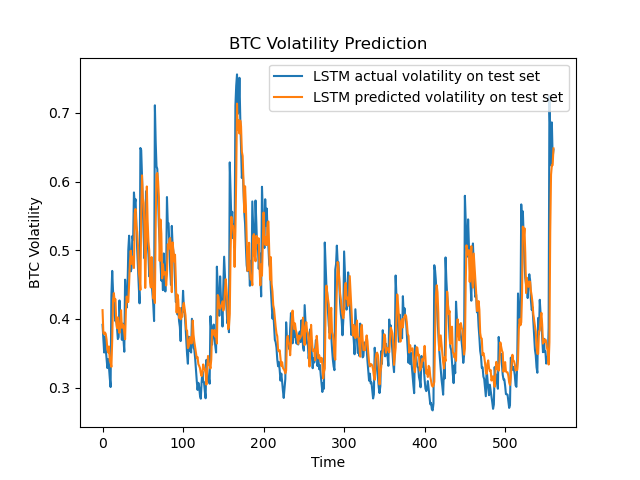}
\caption{Actual Volatility vs LSTM Forecasted Volatility for Test Set}
\label{fig:LSTM_test}
\end{figure} 

\begin{figure}[h!]
    \centering
\includegraphics[scale=0.4]{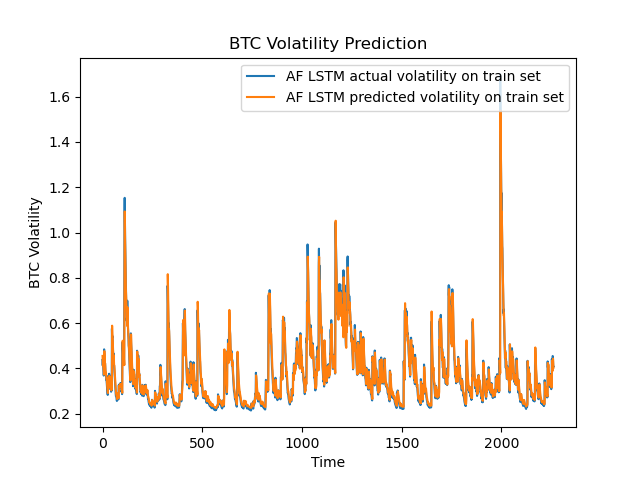}
\caption{Actual Volatility vs AF-LSTM Forecasted Volatility for Train Set}
\label{fig:AF_LSTM_train}
\end{figure}

\begin{figure}[h!]
    \centering
\includegraphics[scale=0.4]{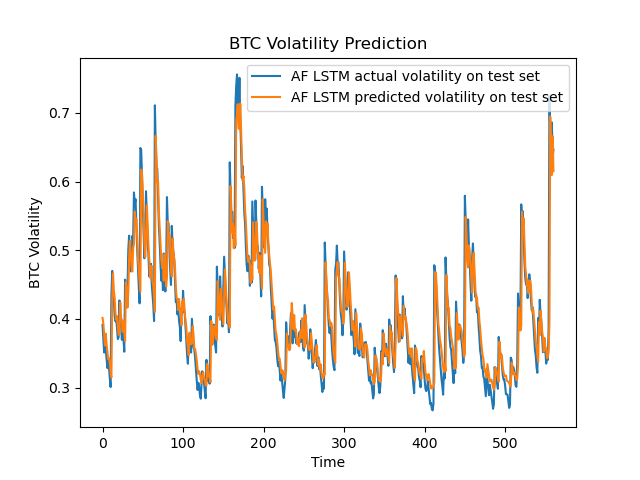}
\caption{Actual Volatility vs AF LSTM Forecasted Volatility for Test Set}
\label{fig:AF_LSTM_test}
\end{figure}

According to the predictions in Figure  \ref{fig:LSTM_train} and Figure  \ref{fig:AF_LSTM_train}, the AF-LSTM is able to catch more patterns than the LSTM. In fact, the lack of amplitude perceived in the LSTM is partially corrected by the AF-LSTM on the train set. When we look at the predictions on the test set in Figure  \ref{fig:LSTM_test} and Figure  \ref{fig:AF_LSTM_test}, the main difference to notice is the reduction of the offset between the LSTM and AF-LSTM results compared to the train set. However, the superiority of the AF-LSTM in approximating the rolling volatility is still demonstrated.

\medskip

The RMSE confirm those results:

\begin{table}[h!]
\centering
\begin{tabular}{ccc}
\hline
Dataset & LSTM RMSE & AF LSTM RMSE \\
\hline
Train Set & 0.061 & 0.051 \\
Test Set & 0.044 & 0.042
\label{Table 1 : Result on train set and test set}
\end{tabular}
\end{table}

\section{Results}

By reducing the time and space complexity of the prediction algorithm, the AF-LSTM Figure \ref{fig:AF_LSTM_test} has a real advantage over the LSTM in terms of sequential computation time on a large data set. In fact, the Attention Free Mechanism has a linear memory complexity with respect to both the context size and the dimension of features, making it compatible to both large input and model sizes.  Moreover, the loss function and predictions are improved as additional information are captured by this model. The AF-LSTM thus demonstrates competitive performance while providing increased efficiency at the same time.

\section{Conclusion}
The GARCH(1,1) derives its value based solely on the most recent updates of $\epsilon$  and $\sigma$, which is helpful if the price changes are relatively similar. However, this model's weakness emerges when there are more sudden jumps in log returns. To consider the repercution of theses shocks on the volatility, the LSTM model should be used, as it tries to mimick the real data with lags. To boost the performance and increase the power of prediction, using an Attention Free LSTM layer can be very useful. It eliminates the quadratic complexity of the self attention mechanism while selecting the past observation that will have a positive impact on the prediction. In fact, instead of carrying out the dot product for the creation of the attention matrix, a weighted average of the values is carried out for each target position. It maintains all the advantages of the dot product without the computation cost.

\bibliography{bib}
\nocite{*}

\end{document}